\documentclass{article}
\usepackage{ijcai21}

\usepackage{times}
\usepackage{soul}
\usepackage{url}
\usepackage[hidelinks]{hyperref}
\usepackage[small]{caption}
\usepackage{graphicx}
\usepackage{amsmath}
\usepackage{amsthm}
\usepackage{booktabs}
\usepackage{algorithm}
\usepackage{algorithmic}
\usepackage{color}
\usepackage{comment}
\usepackage{overpic}
\usepackage{amssymb}

\usepackage{diagbox}
\usepackage{multirow}
\usepackage{arydshln}

\newcommand{\xw}[1]{\textcolor{black}{#1}}

\title{Boundary Knowledge Translation based Reference Semantic Segmentation}

\author{
Lechao Cheng$^{1}$ \footnotemark[2]
\and
Zunlei Feng$^2$ \footnotemark[2]
\and
Xinchao Wang$^3$
\and
Ya Jie Liu $^1$
\and
Jie Lei $^4$
\and
Mingli Song$^2$ \footnotemark[1]
\affiliations
$^1$Zhejiang Lab\\
$^2$Zhejiang University\\
$^3$National University of Singapore\\
$^4$Zhejiang University Of Technology\\
\emails
\{chenglc,liuyj\}@zhejianglab.com,
\{zunleifeng,brooksong\}@zju.edu.cn,
xinchao@nus.edu.sg,
jasonlei@zjut.edu.cn
}

\begin{document}
\def\mathbi#1{\textbf{\em #1}}
\maketitle

\renewcommand{\thefootnote}{\fnsymbol{footnote}}
\footnotetext[1]{Corresponding authors.}
\footnotetext[2]{These authors contributed equally to this work.}

\begin{abstract}
Given a reference object of an unknown type in an image,
human observers can effortlessly find the
objects of the same category  in another image
and precisely tell their visual boundaries.
Such visual cognition capability of humans
seems absent from the current research spectrum
of computer vision.
Existing segmentation networks, for example,
 rely on a humongous amount of labeled data,
which is laborious and costly to collect and annotate;
besides, the performance of segmentation networks
tend to downgrade as the number of the category increases.
In this paper,
we introduce a novel Reference semantic segmentation Network (Ref-Net)
to conduct visual boundary knowledge translation.
Ref-Net contains a Reference Segmentation Module
(RSM) and a Boundary Knowledge Translation Module (BKTM).
Inspired by the human recognition mechanism,
RSM is devised only to segment the same category objects
based on the features of the reference objects.
BKTM, on the other hand, introduces
two boundary discriminator branches
to conduct inner and outer boundary segmentation of the target object
in an adversarial manner,
and translate the annotated boundary knowledge
of open-source datasets into the segmentation network.
Exhaustive experiments demonstrate that, with tens of
finely-grained annotated samples as guidance,
Ref-Net achieves results on par with fully
supervised methods on six datasets. 
\end{abstract}

\section{Introduction}
In recent years, deep neural networks have triumphed over many computer vision problems, including semantic segmentation, which is critical in emerging autonomous driving and medical image diagnostics applications.
In general, training deep neural networks requires a humongous amount of labeled data, which is laborious and costly to collect and annotate.
To alleviate the annotation burden, some learning techniques, such as few-shot learning and transfer learning, have been proposed.
The former aims to train models using only a few annotated samples,
while the latter focuses on transferring the models learned on one domain
to another novel one.
Despite the recent progress  in few-shot and transfer learning,
existing methods are still prone to either inferior results, 
or the rigorous requirement that the two tasks are strongly related 
and a large number of annotated samples are required.

\begin{figure}
\centering
\includegraphics[scale =0.62]{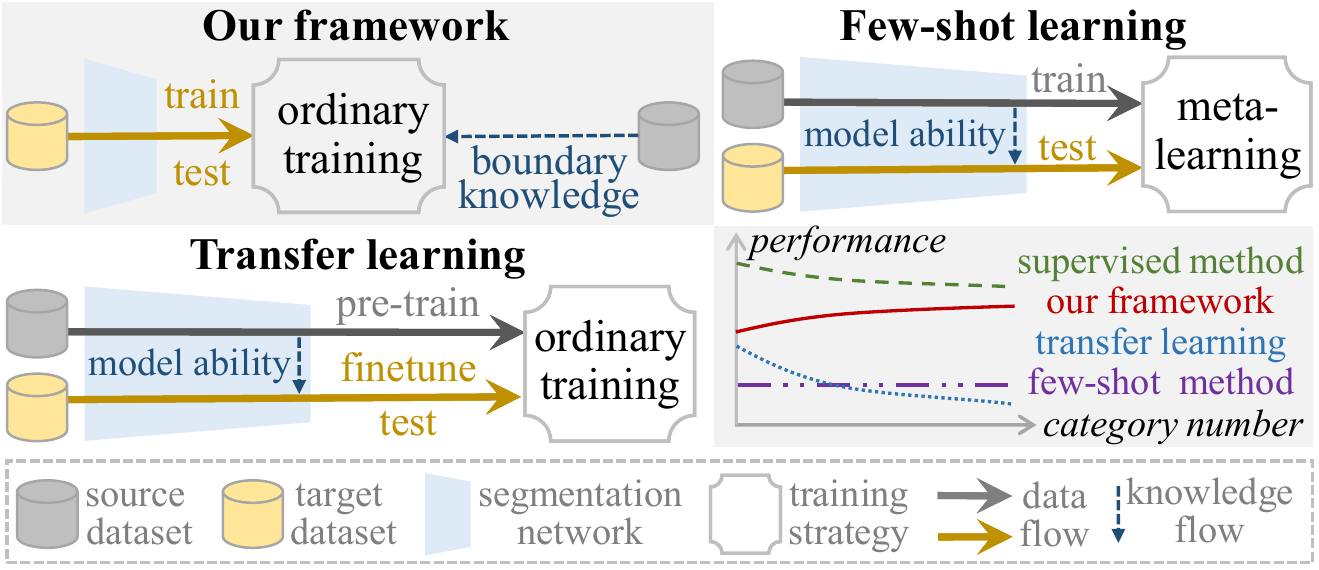}
\caption{The differences between different frameworks.
The main differences include three aspects.
First, the source sample does not pass through the segmentation network in our framework,
\xw{while others' input contains both source and target samples}.
Second, few-shot learning adopts meta-learning as the training strategy, while others not.
Third, our framework translates dataset-level knowledge into the segmentation network,
while others transfer model-level knowledge for target dataset;
this leads to the fact that the performance of our framework enhances as
the number of the category increases.}
\label{fig:overview}
\end{figure}

For many tasks, including few-shot and transfer learning,
the performances of existing approaches, such as segmentation networks,
deteriorate as the number of object categories
increase,
as demonstrated by prior works~\cite{vinyals2016matching,chen2019unsupervised},
and also by our experiments.
The root cause lies in that,
existing approaches are devised to recognize the category-wise features and segment the corresponding objects.
Recently, boundary-aware features have been introduced to enhance the segmentation results, yet their core frameworks still focus on \xw{ classifying category-wise features and segmenting the corresponding objects}.

In the paper, we propose a novel Reference semantic segmentation Network (Ref-Net) based on visual boundary knowledge translation.
In Ref-Net, \xw{only the target samples pass through the segmentation network},
while the boundary knowledge of open-source datasets is translated into the segmentation network in an adversarial manner.
That means only the data flow of the target dataset will pass through the segmentation network.
Fig.~\ref{fig:overview} shows the differences between the proposed framework and related frameworks.
More importantly, Ref-Net is the first proposed dataset-level knowledge translation framework,
which is different from the model-level knowledge transfer of related frameworks.
The fatal limitation of model-level knowledge transfer is
that the specific segmentation ability of the model for
the source categories is beneficial for the segmentation of
target categories, but also will limit the bound of the
performance on the target categories.
The most direct evidence is that the segmentation performance will drop dramatically when the number of the category increases.
By contrast, it is noteworthy that the performance will increase when the number of the category increase for Ref-Net.

\begin{figure*}[!t]
\centering
\vspace{1.5em}
\begin{overpic}[scale =0.74]{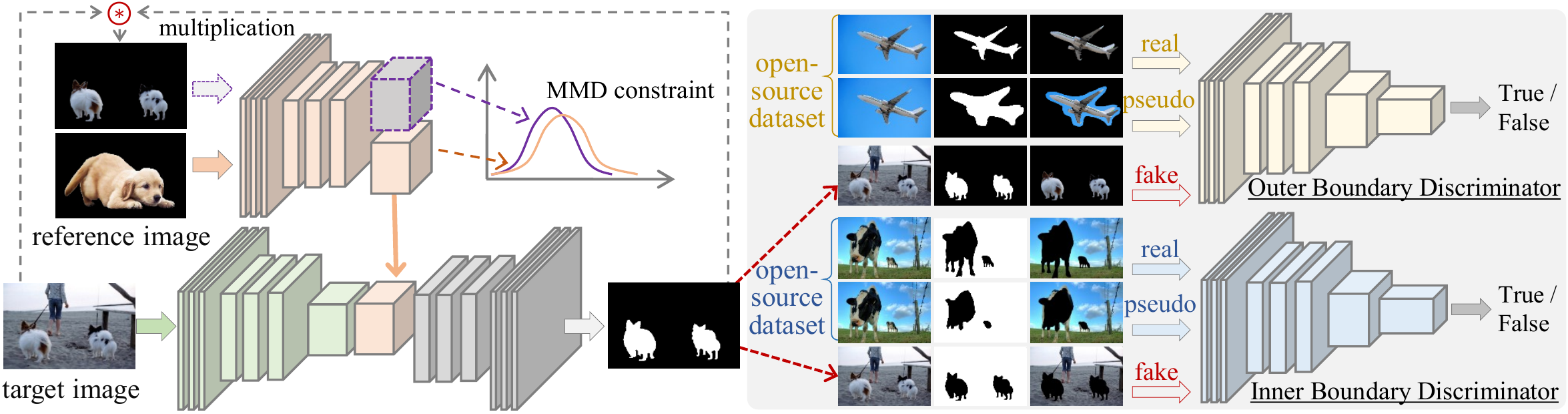}
\put(9.0,26.5){\small Reference Segmentation Module (RSM)}
\put(55.0,26.5){\small Boundary Knowledge Translation Module (BKTM)}
\end{overpic}
\caption{The framework of Ref-Net, which is composed of RSM and BKTM.
In RSM, with the extracted features of reference image as condition,
the segmentation network branch will find and segment the same category objects in the target image.
Meanwhile, the MMD constraint is adopted to constrain the distribution consistency between the representation of the reference object and representation of segmented objects.
The feature extraction branch and the encoder of the segmentation network branch share the same structure and parameters.
The BKT contains the outer boundary discriminator and the inner boundary discriminator,
which are devised for distinguishing whether the segmented objects and background contain residual features.
}
\label{fig:framework}
\end{figure*}

With a reference object,
the human vision system can effortlessly find the same category objects in another image and precisely tell their visual boundary, even if they belong to an unknown category.
Inspired by the above fact, the Ref-Net is devised with a Reference Segmentation Module (RSM) and a Boundary Knowledge Translation Module (BKTM),
as shown in Fig.~\ref{fig:framework}.
In RSM, the reference object guides the segmentation network to {find objects of the same category}.
Meanwhile, BKTM is proposed to assist the segmentation network in
{handling the accurate boundary segmentation},
through translating the annotated boundary knowledge of open-source datasets. \xw{The object category of open-source datasets can be totally different from the target dataset}.

RSM is devised for segmenting the same category objects in the target samples with several annotated reference objects.
For the annotated reference image, the extracted features by the first branch are set as a condition,
which will be concatenated with the target image features.
Based on the condition, the semantic segmentation network learns to find and segment the same category objects in the target images.

To alleviate the burden of laborious and costly annotations,
we adopt BKTM to translate the general boundary knowledge of abundant open-source segmentation datasets into the boundary segmentation ability of segmentation network.
For accurate segmentation, the segmented object should not contain any background feature; meanwhile, the segmented background should not have residual object features.
Inspired by this fact,  BKTM is designed to comprise an inner boundary discriminator and an outer discriminator, as shown in Fig.~\ref{fig:framework}.
The outer boundary discriminator distinguishes whether the segmented objects contain the features of the outer background.
Meanwhile, the inner boundary discriminator distinguishes whether the segmented background contains features of the inner objects.


Our contribution is therefore the first dataset-level knowledge translation based Ref-Net for semantic segmentation,
which brings increased performance as the number of target category increases.
Also, a boundary-aware self-supervision and a category-wise constraint are proposed to enhance the segmentation consistency
on both the image-level and representation level, respectively.
We evaluate the proposed Ref-Net on a wide domain of image datasets,
and show that, with only ten annotated samples, our method achieves close results on par with fully supervised ones.

\section{Related Works}
For \textbf{boundary-aware semantic segmentation}, the commonly adopted framework is a two-branch network that simultaneously predicts segmentation maps and boundaries~\cite{takikawa2019gated-scnn:}.
Unlike predicting the boundary directly,
some strategies, such as pixel's distance to boundary~\cite{hayder2017boundary-aware}, boundary-aware filtering~\cite{khoreva2017simple},
boundary refinement~\cite{zhang2017global-residual}, boundary weights~\cite{qin2019basnet:}, are proposed for improving the segmentation performance on boundary.
Unlike the above methods, the proposed BKTM focuses on the outer boundary of the object and inner boundary of the background.
Meanwhile,
two boundary discriminators are devised for discriminating whether the outer boundary of the object and inner boundary of the background contain residual features.

\textbf{GAN based segmentation} contains two categories: mask distribution-based methods ~\cite{arbelle2018microscopy} and composition fidelity based methods~\cite{chen2019unsupervised}.
The former discriminates mask distribution between the predicted mask and GT mask, while the latter adopts discriminator to discriminate fidelity of the composite images.
Unlike the above GAN-based methods,
an adversarial strategy is adopted to translate the annotated boundary knowledge of the open-source dataset into the segmentation network by two boundary discriminators.

\textbf{Few-shot Segmentation (FSS)} aims at training a segmentation network that can segment new category well with only a few labeled samples of those new category. It contains parametric matching-based methods
\cite{zhang2018sg-one:,xian2019semantic}, prototype-based methods~\cite{wang2019panet:}, GCN-based methods~\cite{LiuAAAI21}, R-CNN based methods~\cite{yan2019meta} and proposal-free based methods~\cite{gao2019ssap:}.
Ref-Net is expected to gain a general segmentation ability and focus on the segmentation ability of the target categories.
In addition, the category number of support category in FSS is usually larger than two,
while Ref-Net can handle a single category source dataset.

\textbf{Transfer learning based segmentation} contains pseudo-sample generation methods~\cite{han2017transferring}, iterative optimization methods~\cite{zou2018unsupervised},
graph-based methods~\cite{YangNeurIPS20},
and distillation methods~\cite{yangCVPR20}.
Those methods aim at transferring the models' ability on source dataset into target datasets, where the capacity for source datasets still occupy the part of the segmentation ability of the model. However, the Ref-Net aims at learning a general segmentation ability with a reference image as a condition. Training on target datasets will bring more focused segmentation ability on the target categories.
In addition, transfer learning based methods require the two domains as similar as possible, but Ref-Net has no such requirement.


\section{Proposed Method}

The proposed Ref-Net is composed of a Reference Segmentation Module (RSM) and a Boundary Knowledge Translation Module (BKTM), as shown in Fig.~\ref{fig:framework}.
RSM is designed for  segmenting the same category objects in the target samples with the reference image as condition,
which is inspired by the human recognition mechanism with a reference object.
BKTM contains two boundary discriminators,
which are devised for distinguishing
whether segmented objects contain outer features of the background and segmented background contains inner features objects, respectively.
BKTM aims to translate the annotated boundary knowledge of open-source datasets into the ability of the segmentation network,
which will dramatically reduce the requirement of labeled samples for the target categories.

Formally, given a target dataset $\mathbb{S}$ with object category labels,
a reference image dataset $\mathbf{S}$ with fine-grained object mask annotation
and an open-source dataset $\mathbf{O}$ with fine-grained object mask annotation,
the goal of Ref-Net is to learn a segmentation network for $\mathbb{S}$ with $\mathbf{S}$ as condition and guidance,
while BKTM translates the annotated boundary knowledge of $\mathbf{O}$ into the segmentation network.
It is noticed that the reference image dataset $\mathbf{S}$ are chosen from the target dataset $\mathbb{S}$,
which has the same object category.
The object category of open-source dataset $\mathbf{O}$ can be totally different from the category of the target dataset $\mathbb{S}$.


\subsection{Reference Segmentation Module}
Inspired by the human recognition mechanism with a reference object,
RSM is devised to be composed of two network branches: the reference feature extraction branch and the target segmentation branch.
The target segmentation branch is designed to be an encoder-decoder architecture.
To maintain the consistency of the feature space,
the reference feature extraction branch $\mathcal{F}^{e}_{\theta}$ has the same network architecture and parameter as the encoder of the target segmentation branch.

Give a target image $\overline{\mathbi{x}}\in \mathbb{S}$ and a reference image $(\mathbi{x}^c_k,\mathbi{m}^c_k) \in \mathbf{S}$,
the extracted representations can be denoted as
$\overline{\mathbi{r}}=\mathcal{F}^e_{\theta}(\overline{\mathbi{x}})$
and $\mathbi{r}^c_k=\mathcal{F}^e_{\theta}(\mathbi{x}^c_k*\mathbi{m}^c_k)$, where the $*$ denotes pixel-wise multiplication.
With the concatenated representation $[\overline{\mathbi{r}}, \mathbi{r}^c_k]$,
the decoder $\mathcal{F}^d_{\theta}$ of the  target segmentation branch predicts the mask $\tilde{\mathbi{m}}=\mathcal{F}^d_{\theta}([\overline{\mathbi{r}}, \mathbi{r}^c_k])$.
For simplicity, the reference segmentation is formulated as
$\tilde{\mathbi{m}}=\mathcal{F}_{\theta}(\overline{\mathbi{x}},\mathbi{x}^c_k*\mathbi{m}^c_k)$,
where the $\mathcal{F}_{\theta}$ denotes the two-branch segmentation network.

\textbf{Limited Sample Supervision.} In the training stage,
the limited annotated samples are also fed to the target segmentation branch,
which will generate direct supervision information.
So, given a random image $(\mathbi{x},\mathbi{m}) \in \mathbf{S}$ and a reference image $(\mathbi{x}^c_k,\mathbi{m}^c_k) \in \mathbf{S}$,
the segmented result $\tilde{\mathbi{m}}=\mathcal{F}_{\theta}(\mathbi{x},\mathbi{m}^c_k*\mathbi{x}^c_k)$ is expected to approximate the GT mask $\mathbi{m}$,
which can be achieved by minimizing the pixel-wise two-class Dice loss $\mathbf{\mathcal{L}_{dic}}$:
\begin{equation}\label{eq1}
\mathbf{\mathcal{L}_{dic}}=1-\frac{2|\tilde{\mathbi{m}} \cap \mathbi{m}|+\tau}{|\tilde{\mathbi{m}}|+|\mathbi{m}|+\tau},
\end{equation}
where, $\tau$ is the Laplace smoothing parameter for preventing zero error and reducing overfitting,
the GT mask $\mathbi{m}$ will be an all-zero mask when the random image $\mathbi{x}$ and reference image $\mathbi{x}^c_k$ has no same category object.
For the limited annotated samples, the data augmentation strategy is adopted to increase the samples' diversity.

\textbf{Representation Consistency Constraint.}
The same category objects in the reference image and target image should have a similar representation distribution.
So, we adopt the Maximum Mean Discrepancy ($\mathcal{MMD}$) to constrain the representation distribution consistency.
With the reference image $(\mathbi{x}^c_k,\mathbi{m}^c_k) \in \mathbf{S}$ and the same category target image $\overline{\mathbi{x}}$,
the predicted mask is $\tilde{\mathbi{m}}=\mathcal{F}_{\theta}(\overline{\mathbi{x}},\mathbi{m}^c_k*\mathbi{x}^c_k)$.
The representation consistency loss $\mathbf{\mathcal{L}_{rep}}$ is defined as follows:
\begin{equation}\label{eq2}
\mathbf{\mathcal{L}_{rep}}=\mathcal{MMD}\{f(\tilde{\mathbi{m}}*\overline{\mathbi{x}}),f(\mathbi{m}^c_k*\mathbi{x}^c_k)\},
\end{equation}
where $f$ is the feature encoder for reference image.


\textbf{Boundary-aware Self-supervision.}
To reduce the number of labeled samples,
inspired by ~\cite{wang2019self-supervised}, we propose a boundary-aware self-supervision strategy,
which can strengthen the boundary consistency of target objects.
The core idea is that for the same network, the predicted mask of the \emph{transformed input image} should be equal to the \emph{transformed  mask} predicted by the network with the original image as input.
 Formally, for the robust segmentation network, given an affine transformation matrix $\mathbi{A}$,
 the segmented result $\tilde{\mathbi{m}}'=\mathcal{F}_{\theta}(\mathbi{A}\overline{\mathbi{x}},\mathbi{x}^c_k*\mathbi{m}^c_k)$
 of the transformed image $\mathbi{A}\overline{\mathbi{x}}$ and
 the transformed result  $\mathbi{A}\mathcal{F}_{\theta}(\overline{\mathbi{x}},\mathbi{x}^c_k*\mathbi{m}^c_k)$
 should be consistent in the following way:
 $\mathcal{F}_{\theta}(\mathbi{A}\overline{\mathbi{x}},\mathbi{x}^c_k*\mathbi{m}^c_k)=\mathbi{A}\mathcal{F}_{\theta}(\overline{\mathbi{x}},\mathbi{x}^c_k*\mathbi{m}^c_k)$.
 Furthermore, we obtain the boundary neighborhood weight map $\mathbi{w}'$ and $\mathbi{w}$ as follows:
\begin{equation}\label{eq2}
\begin{split}
\mathbi{w}' = & \mathfrak{D}_r(\tilde{\mathbi{m}}')-\mathfrak{E}_r(\tilde{\mathbi{m}}'), \\
\mathbi{w} = & \mathfrak{D}_r(\tilde{\mathbi{m}})-\mathfrak{E}_r(\tilde{\mathbi{m}}),
\end{split}
\end{equation}
where, $\mathfrak{D}_r$ and $\mathfrak{E}_r$ denote the dilation and erosion operation with a disk strel of radius $r$, respectively.
The weight map $\mathbi{w}'$ and $\mathbi{w}$ can strengthen the boundary consistency.
The boundary-aware self-supervision loss $\mathbf{\mathcal{L}_{sel}}$ is defined as follows:
\begin{equation}\label{eq3}
\mathbf{\mathcal{L}_{sel}}=||\mathbi{w}'\!*\mathcal{F}_{\theta}(\mathbi{A}\overline{\mathbi{x}},\mathbi{x}^c_k*\mathbi{m}^c_k)
- \mathbi{A}\{\mathbi{w}\!*\mathcal{F}_{\theta}(\overline{\mathbi{x}},\mathbi{x}^c_k*\mathbi{m}^c_k)\}||^2_2,
\end{equation}
where, $\mathbi{w}'$ and $\mathbi{w}$ are the weight maps of the predict masks
 $\tilde{\mathbi{m}}'$ and $\tilde{\mathbi{m}}$, respectively. The boundary-aware self-supervision mechanism not only strengthens the boundary consistency but also eliminates the unreasonable holes.

\begin{table*}[!t]
\resizebox{\textwidth}{!}{
\centering
\begin{tabular}{ccccccccccccccccccc}
\toprule
\textbf{Type ($\mathbb{S} \leftarrow \mathbf{O}$)}
    & \multicolumn{3}{c}{$\mathcal{P.}$(\footnotesize{\emph{Cityscapes}} $\leftarrow$ \footnotesize{\emph{SYNTHIA}}  ) }
    & \multicolumn{6}{c}{\emph{Multiple Category} (\emph{Half-category} $\leftarrow$ \emph{Half-category}})
    & \multicolumn{9}{c}{\emph{Single Category} (\emph{One} $\leftarrow$ \emph{MixAll}$^{-}$)} \\
    \cmidrule(r){1-1} \cmidrule(r){2-4}  \cmidrule(r){5-10} \cmidrule(r){11-19}
\textbf{Dataset}
    & \multicolumn{3}{c}{\textbf{Cityscapes} }
    & \multicolumn{3}{c}{\textbf{SBD}}
    & \multicolumn{3}{c}{\textbf{THUR}}
    & \multicolumn{3}{c}{\textbf{Bird}}
    & \multicolumn{3}{c}{\textbf{Human}}
    & \multicolumn{3}{c}{\textbf{Flower}}\\
    \cmidrule(r){1-1}
     \cmidrule(r){2-4}  \cmidrule(r){5-7} \cmidrule(r){8-10} \cmidrule(r){11-13}  \cmidrule(r){14-16} \cmidrule(r){17-19}
\footnotesize{\textbf{Method}$\backslash$\textbf{Index}}       &\footnotesize{\textbf{MPA}}&\footnotesize{\textbf{MIoU}}&\footnotesize{\textbf{FWIoU}}   &\footnotesize{\textbf{MPA}}&\footnotesize{\textbf{MIoU}}&\footnotesize{\textbf{FWIoU}}  &\footnotesize{\textbf{MPA}}&\footnotesize{\textbf{MIoU}}&\footnotesize{\textbf{FWIoU}}  &\footnotesize{\textbf{MPA}}&\footnotesize{\textbf{MIoU}}&\footnotesize{\textbf{FWIoU}}   &\footnotesize{\textbf{MPA}}&\footnotesize{\textbf{MIoU}}&\footnotesize{\textbf{FWIoU}}   &\footnotesize{\textbf{MPA}}&\footnotesize{\textbf{MIoU}}&\footnotesize{\textbf{FWIoU}}\\
\cmidrule(r){1-1} \cmidrule(r){2-4}  \cmidrule(r){5-7} \cmidrule(r){8-10} \cmidrule(r){11-13}  \cmidrule(r){14-16} \cmidrule(r){17-19}
\textbf{CAC}        &-- &-- & --     &-- & -- & --   &- &- & -      &48.28 &26.42 & 27.06      &47.14 &24.28 &23.72        &36.96 &24.24 &35.34\\
\textbf{ReDO}       &-- &-- & --     &-- &-- & --  &- &- & -    &50.00 &38.53 & 33.01     &50.00 &35.72 & 36.49      &70.00 &58.58 &43.82 \\
\cmidrule(r){1-1} \cmidrule(r){2-4}  \cmidrule(r){5-7} \cmidrule(r){8-10} \cmidrule(r){11-13}  \cmidrule(r){14-16} \cmidrule(r){17-19}
\textbf{SG-One}       &-- &-- & --      &32.61 &24.91 & 47.30    &84.11 &71.51 & 69.27      &78.66 &61.43 & 74.68     &72.46 &56.72 & 56.93       &87.05 &74.73 &76.77\\
\textbf{PANet}        &-- &-- & --      &31.03 &20.09 & 49.78    &66.90 &55.75 & 84.39      &66.94 &57.83 & 69.42     &76.49 &60.54 & 63.29     &68.43 &69.25 &69.94 \\
\textbf{SPNet}        &-- &-- & --      &30.10 &21.92 & 39.09    &84.11 &71.51 & 69.27    &79.65 &76.92 & 78.01      &75.90 &60.42 & 62.90     &88.43 &79.21 & 80.43\\
\textbf{CANet}       &-- &-- & --      &40.17 &32.02 & 40.82    &59.51 &50.49 & 79.02    &85.36 &76.02 & 85.01      &95.29 &90.98 & 90.99     &81.68 &70.83 & 73.96\\
\cmidrule(r){1-1} \cmidrule(r){2-4}  \cmidrule(r){5-7} \cmidrule(r){8-10} \cmidrule(r){11-13}  \cmidrule(r){14-16} \cmidrule(r){17-19}
\textbf{ALSSS}      &50.62 &41.19 & \textcolor{blue}{77.53}    &33.45 &23.86 & 54.95    &80.04 &60.28 & 77.63   &51.54 &39.48 & 66.09     &76.26 &60.42 & 63.90       &79.32 &85.21 &87.32\\
\textbf{USSS}       &51.36 &40.68 & 71.99    &55.26 &41.07 & 68.53     &84.12 &71.51 & 84.39   &49.64 &41.27 & 67.95     &77.75 &62.40 & 62.28      &95.15 &90.81 &91.36\\
\cmidrule(r){1-1} \cmidrule(r){2-4}  \cmidrule(r){5-7} \cmidrule(r){8-10} \cmidrule(r){11-13}  \cmidrule(r){14-16} \cmidrule(r){17-19}
\textbf{Trans.(10)}        &37.99 &30.92 & 70.22     &20.75 &14.00 & 70.22     &74.09 &62.25 & 87.73   &79.83 &65.07 & 74.71     &85.44 &75.71 & 75.72        &82.69 &78.66 &76.39\\
\textbf{Trans.(100)}       &46.65 &37.57 & 71.75     &31.67 &23.71 & 71.25     &84.37 &\textcolor[rgb]{0.80,0.00,0.00}{75.28} & \textcolor[rgb]{0.80,0.00,0.00}{91.92}   &\textcolor[rgb]{0.80,0.00,0.00}{91.56} & \textcolor[rgb]{0.80,0.00,0.00}{83.23} & 84.06    &95.00 &90.46 & 90.48      &89.29 &81.33 &89.05\\
\cmidrule(r){1-1} \cmidrule(r){2-4}  \cmidrule(r){5-7} \cmidrule(r){8-10} \cmidrule(r){11-13}  \cmidrule(r){14-16} \cmidrule(r){17-19}
\textbf{Gated-SCNN}    &52.89 &39.37 & 71.45     &49.44 &38.44 & 84.74     &91.11 &78.77 & 90.41      &\textcolor[rgb]{0.00,0.50,0.00}{94.90} &\textcolor[rgb]{0.00,0.50,0.00}{90.71} & \textcolor[rgb]{0.00,0.50,0.00}{96.40}      &\textcolor[rgb]{0.00,0.50,0.00}{98.82} &\textcolor[rgb]{0.00,0.50,0.00}{97.61} & \textcolor[rgb]{0.00,0.50,0.00}{97.60}     &95.61 &92.83 & 93.92\\
\textbf{BFP}       &51.03 &36.43 & 70.20     &50.24 &42.31 & 45.20     &75.02 &77.32 & 84.25      &92.83 &87.48 & 91.45      &97.81 &96.30 & 96.41     &95.44 &92.30 & 93.81\\
\cdashline{1-1}[0.8pt/2pt] \cdashline{2-4}[0.8pt/2pt]  \cdashline{5-7}[0.8pt/2pt] \cdashline{8-10}[0.8pt/2pt] \cdashline{11-13}[0.8pt/2pt]  \cdashline{14-16}[0.8pt/2pt] \cdashline{17-19}[0.8pt/2pt]
\textbf{Unet}        &52.80 &42.98 & 74.66     &73.48 & 61.45 & 89.17     &91.04 &\textcolor{blue}{84.53} & \textcolor{blue}{94.85}     &91.88 &86.41  &92.06     &97.88 &95.86 &95.87        &96.39 &93.44 & 93.89 \\
\textbf{FPN}         &55.10 &45.12 & 74.38     &72.78 &62.24 & 89.16     &88.78 &81.80 & 93.91    &92.86 &86.53  &92.06     &98.19 &96.45 &96.46         &\textcolor[rgb]{0.00,0.50,0.00}{97.16} &\textcolor{blue}{94.22} & \textcolor{blue}{94.59} \\
\textbf{LinkNet}     &43.99 &35.02 & 73.96     &\textcolor{blue}{74.33} &62.37 & \textcolor{blue}{89.31}     &90.74 &83.66 & 94.52     &93.04 &86.03  &91.77     &97.41 &94.97 &94.98       &96.82 &\textcolor[rgb]{0.00,0.50,0.00}{94.26} & \textcolor[rgb]{0.00,0.50,0.00}{94.65}\\
\textbf{PSPNet}      &39.76 &32.70 & 69.01     &49.23 &39.65 &82.43     &82.22 &73.13 & 90.95   &87.01 &79.47  &87.97     &97.02 &94.22 &94.23        &95.77 &91.46 & 91.99\\
\textbf{PAN}         &55.60 &45.13 & 74.17     &72.30 &60.43 & 88.74     &90.75 &81.61 & 93.76   &93.86 & 87.07 &92.38     &98.15 &96.37 &96.38        &96.71 &93.64 & 94.06\\
\textbf{DeeplabV3+}  &\textcolor{blue}{57.49} &\textcolor{blue}{46.45} & 75.19     &74.32 & \textcolor{blue}{63.06} & \textcolor[rgb]{0.00,0.50,0.00}{89.38}     &\textcolor[rgb]{0.00,0.50,0.00}{92.48} &\textcolor[rgb]{0.00,0.50,0.00}{84.92} & \textcolor[rgb]{0.00,0.50,0.00}{94.94}   &\textcolor{blue}{94.88} &\textcolor{blue}{89.62}  &93.95     &\textcolor{blue}{98.28} &\textcolor{blue}{96.62}& \textcolor{blue}{96.63}       &96.65 &93.80 & 94.23\\
\cmidrule(r){1-1} \cmidrule(r){2-4}  \cmidrule(r){5-7} \cmidrule(r){8-10} \cmidrule(r){11-13}  \cmidrule(r){14-16} \cmidrule(r){17-19}
\mathbi{R}(0)     &-- &-- & --  &-- &-- & --      &-- &-- & --    &86.02 &70.69 & 85.24      &69.95 &53.03 & 53.00     &82.88 &71.89 & 75.13\\
\mathbi{R}(10)    &\textcolor[rgb]{0.80,0.00,0.00}{53.44} &\textcolor[rgb]{0.80,0.00,0.00}{45.01} & \textcolor[rgb]{0.80,0.00,0.00}{81.61}     &\textcolor[rgb]{0.80,0.00,0.00}{75.64} &\textcolor[rgb]{0.80,0.00,0.00}{62.10} & \textcolor[rgb]{0.80,0.00,0.00}{71.31}     &\textcolor[rgb]{0.80,0.00,0.00}{88.42} &74.84 & 88.78      &90.85 &77.14 & \textcolor[rgb]{0.80,0.00,0.00}{89.34}      &\textcolor[rgb]{0.80,0.00,0.00}{95.86} &\textcolor[rgb]{0.80,0.00,0.00}{92.06} & \textcolor[rgb]{0.80,0.00,0.00}{92.07}   &\textcolor[rgb]{0.80,0.00,0.00}{95.86} &\textcolor[rgb]{0.80,0.00,0.00}{91.02} & \textcolor[rgb]{0.80,0.00,0.00}{92.09}\\
\mathbi{R}(fully)  &\textcolor[rgb]{0.00,0.50,0.00}{58.21} &\textcolor[rgb]{0.00,0.50,0.00}{46.91} & \textcolor[rgb]{0.00,0.50,0.00}{86.34}     &\textcolor[rgb]{0.00,0.50,0.00}{79.14} &\textcolor[rgb]{0.00,0.50,0.00}{63.07} & 71.68     &\textcolor{blue}{92.19} &76.84 & 90.01   &94.76 &87.27 & \textcolor{blue}{94.15}     &97.14 &94.44 & 94.44       &\textcolor{blue}{97.03} &92.05 & 92.94\\
\bottomrule
\end{tabular}
}
\caption{The performance comparison of different methods.
`$\mathbb{S} \leftarrow \mathbf{O}$' denotes knowledge of open-source dataset $\mathbf{O}$ are translated into the network for target dataset $\mathbb{S}$.
`--' denotes that the dateset is not suitable for the method.
`$\mathcal{P.}$' denotes the  panoramic.
`\emph{Half-category}' denotes half category of each dataset.
`\emph{MixAll}$^{-}$' denotes mix category open-source dataset without the target category.
`Trans.(K)' denotes transfer learning method with K labeled samples.
\textcolor[rgb]{0.00,0.50,0.00}{Green} and
\textcolor{blue}{Blue} indicate the best and second-best performance among all methods.
\textcolor[rgb]{0.80,0.00,0.00}{Red} indicates the best performance among all non-fully supervised methods.
$\mathbi{R}(K)$ denotes the Ref-Net with $K$ labeled samples (All scores in \%).
}
\label{comparing_SOTA}
\end{table*}

\subsection{Boundary Knowledge Translation Module}
Inspired by the fact that humans can segment the boundary of the object through distinguishing whether the inner and outer of boundary contain redundant features,
we devise two boundary discriminators, which can translate the boundary knowledge of the open-source dataset $\mathbf{O}$ into the reference segmentation network.

\textbf{Outer Boundary Discriminator.}
Randomly sampling a pair of samples $\overline{\mathbi{x}}^c \in \mathbb{S}$ and  $(\mathbi{x}^c_k,\mathbi{m}^c_k) \in \mathbf{S}$
 from the target image set $\mathbb{S}$ and the reference image set $\mathbf{S}$.
Next, the segmentation network predicts the mask
$\tilde{\mathbi{m}}=\mathcal{F}_{\theta}(\overline{\mathbi{x}}^c,\mathbi{x}^c_k*\mathbi{m}^c_k)$.
Then,
the segmented objects $\overline{\mathbi{x}}^c_o$ are computed by:
 $\overline{\mathbi{x}}^c_o=\tilde{\mathbi{m}}*\overline{\mathbi{x}}^c$.
 The concatenated triplet $\mathbi{I}^o_a=[\overline{\mathbi{x}}^c,\tilde{\mathbi{m}},\overline{\mathbi{x}}^c_o]$
  is fed to the outer boundary discriminator $\mathcal{D}^o_{\phi}$,
  which discriminates whether the segmented objects $\overline{\mathbi{x}}^c_o$ contain the outer features of the background.
  In the paper, $\mathbi{I}^o_a$ is regarded as a fake triplet.
Meanwhile, choosing an annotated sample $(\underline{\mathbi{x}},\underline{\mathbi{m}})\in \mathbf{O}$ from the open-source dataset $\mathbf{O}$, 
the corresponding $\mathbi{I}^o_e=[\underline{\mathbi{x}},\underline{\mathbi{m}},\underline{\mathbi{x}}_o]$ is labeled as a real triplet.
Furthermore,
we reprocess the GT mask $\underline{\mathbi{m}}$ of samples $\underline{\mathbi{x}}$
by dilation operation $\mathfrak{D}_r$ and get the pseudo triplet
$\mathbi{I}^o_s=[\underline{\mathbi{x}},\mathfrak{D}_r(\underline{\mathbi{m}}),\mathbi{\underline{x}}^{\mathfrak{D}}_o]$,
where $ \underline{\mathbi{x}}\in \mathbf{O}$ and $\underline{\mathbi{x}}^{\mathfrak{D}}_o=\mathfrak{D}_r(\mathbi{\underline{m}})*\underline{\mathbi{x}}$.
The generated pseudo triplet $\mathbi{I}^o_s$ will assist the outer boundary discriminator in distinguishing the outer features of the background.
The adversarial optimization between the segmentation network and outer boundary discriminator will translate the outer boundary knowledge of the source dataset $\mathbf{O}$ into the segmentation network with the following outer boundary adversarial loss $\mathbf{\mathcal{L}^{out}_{adv}}$:
\begin{equation}\label{eq4}
\begin{split}
\mathbf{\mathcal{L}^{out}_{adv}}\!\!= & \frac{1}{2}\!\mathop{\mathbb{E}}\limits_{\mathbi{I}^o_a \sim \mathbb{P}^o_a}\![\mathcal{D}^o_{\phi}\!(\mathbi{I}^o_a)]\!+\!
\frac{1}{2}\!\mathop{\mathbb{E}}\limits_{\mathbi{I}^o_s \sim \mathbb{P}^o_s} [\mathcal{D}^{o}_{\phi}\!(\mathbi{I}^o_s)]\! -\!\!\!\!\mathop{\mathbb{E}}\limits_{\mathbi{I}^o_e \sim \mathbb{P}^o_e}\![\mathcal{D}^{o}_{\phi}\!(\mathbi{I}^o_e)] \\
& +\lambda \! \mathop{\mathbb{E}}\limits_{\mathbi{I}^o\sim \mathbb{P}_{\mathbi{I}^o}}\!\![(\|\nabla_{\mathbi{I}^o} \mathcal{D}^{o}_{\phi}(\mathbi{I}^o)\|_{2}\!-\!1)^2],
\end{split}
\end{equation}
where, the $ \mathbb{P}^o_a$, $\mathbb{P}^o_s$, $\mathbb{P}^o_e$ are the segmented outer boundary distribution, pseudo outer boundary distribution, and real outer boundary distribution, respectively.
The $\mathbb{P}_{\mathbi{I}^o}$ is sampled uniformly along straight lines between pairs of points sampled from the distribution $\mathbb{P}^o_e$ and the segmentation network distribution $ \mathbb{P}^o_a$.
The $\mathbi{I}^o=\varepsilon \mathbi{I}^o_e+(1-\varepsilon)\mathbi{I}^o_a$, where the $\varepsilon$ is a random number between $0$ and $1$.
The gradient penalty term is firstly proposed in WGAN-GP~\cite{gulrajani2017improved}. The $\lambda$ is the gradient penalty coefficient.

\textbf{Inner Boundary Discriminator.}
The inner boundary discriminator
$\mathcal{D}^i_{\varphi}$ is devised for discriminating whether the segmented
background contains the inner features of the object.
To obtain the segmented background, the predict background mask $\tilde{\mathbi{m}}'$ and GT mask $\mathbi{m}'$
are reprocessed with the Not-operation as follows: $\tilde{\mathbi{m}}'=[\mathbf{1}]-\tilde{\mathbi{m}}$,
$\mathbi{m}'=[\mathbf{1}]-\mathbi{m}$, where the $[\mathbf{1}]$
denotes the unit matrix of $\mathbi{m}$'s size.
Then, the corresponding fake triplet $\mathbi{I}^i_a=[\overline{\mathbi{x}},\tilde{\mathbi{m}}',\overline{\mathbi{x}}_b]$,
real triplet $\mathbi{I}^i_e=[\underline{\mathbi{x}},\mathbi{\underline{m}}',\underline{\mathbi{x}}_b]$
and pseudo triplet $\mathbi{I}^i_s=[\underline{\mathbi{x}},\mathfrak{D}_r(\mathbi{\underline{m}}'),\underline{\mathbi{x}}^{\mathfrak{D}}_b]$
are computed in the same manner as done in the outer boundary discriminator.
The generated pseudo triplet $\mathbi{I}^i_s$ will also assist the inner boundary discriminator in distinguishing the inner features of objects.
Similarly, the inner boundary adversarial loss $\mathbf{\mathcal{L}^{in}_{adv}}$ is defined as follows:

\begin{equation}\label{eq5}
\begin{split}
\mathbf{\mathcal{L}^{in}_{adv}}\!\!= & \frac{1}{2}\!\mathop{\mathbb{E}}\limits_{\mathbi{I}^i_a \sim \mathbb{P}^i_a}\![\mathcal{D}^i_{\varphi}\!(\mathbi{I}^i_a)]\!+\!
\frac{1}{2}\!\mathop{\mathbb{E}}\limits_{\mathbi{I}^i_s \sim \mathbb{P}^i_s} [\mathcal{D}^i_{\varphi}\!(\mathbi{I}^i_s)]\!-\!\!\!\!\mathop{\mathbb{E}}\limits_{\mathbi{I}^i_e \sim \mathbb{P}^i_e}\![\mathcal{D}^i_{\varphi}\!(\mathbi{I}^i_e)]\\
& + \lambda \!\mathop{\mathbb{E}}\limits_{\mathbi{I}^i\sim \mathbb{P}_{\mathbi{I}^i}}\!\![(\|\nabla_{\mathbi{I}^i} \mathcal{D}^i_{\varphi}(\mathbi{I}^i)\|_{2}\!-\!1)^2],
\end{split}
\end{equation}
where, the $ \mathbb{P}^i_a$, $\mathbb{P}^i_s$, $\mathbb{P}^i_e$ are the segmented inner boundary distribution, pseudo inner boundary distribution,
and real inner boundary distribution.
$\mathbi{I}^i=\varepsilon \mathbi{I}^i_e+(1-\varepsilon)\mathbi{I}^i_a$.
The optimization on $\mathbf{\mathcal{L}^{in}_{adv}}$ will translate the outer boundary knowledge of the open-source dataset into the segmentation network.

\subsection{Complete Algorithm}
To sum up,
two boundary adversarial losses $\mathbf{\mathcal{L}^{out}_{adv}}$ and $\mathbf{\mathcal{L}^{in}_{adv}}$
are used to translate the visual boundary knowledge of the source dataset $\mathbf{O}$ into the segmentation network.
The basic reconstruction loss $\mathbf{\mathcal{L}_{rec}}$ is adopted to
supervise the segmentation on tens of finely-grained annotated samples of reference image dataset $\mathbb{S}$.
The representation consistency loss $\mathbf{\mathcal{L}_{rep}}$ and the self-supervision loss $\mathbf{\mathcal{L}_{sel}}$
are devised for strengthening the category-wise representation consistency
and the boundary-aware segmentation consistency on target datasets $\mathbb{S}$.
During training, we alternatively optimize the segmentation network $\mathcal{F}_{\theta}$
and two boundary discriminators $ \mathcal{D}^o_{\phi},\mathcal{D}^i_{\varphi}$
using the randomly sampled samples from the reference image dataset $\mathbi{S}$,
target dataset $\mathbb{S}$ and the open-source dataset $\mathbf{O}$, respectively.
For training the segmentation network $\mathcal{F}_{\theta}$, the total loss  $\mathbf{\mathcal{L}_{seg}}$ is calculated as follows:
\begin{equation}\label{eq6}
\mathbf{\mathcal{L}_{seg}}= \xi \mathbf{\mathcal{L}_{dic}}+ \zeta \mathbf{\mathcal{L}_{rep}} +\eta \mathbf{\mathcal{L}_{sel}}
-\mathcal{D}^{o}_{\phi}(\mathbi{I}^o_a)-\mathcal{D}^{i}_{\varphi}(\mathbi{I}^i_a).
\end{equation}
For training the inner and outer discriminators, the Eq.(\ref{eq4}) and Eq.(\ref{eq5}) are adopted.

\section{Experiments}

\textbf{Dataset.}
The target datasets we adopted contain Cityscapes, SBD, THUR, Bird, Flower, Human.
What's more, the open-source datasets (MSRA10K, MSRA-B, CSSD, ECSSD, DUT-OMRON, PASCAL-Context, HKU-IS, SOD, SIP1K)
are merged into \emph{MixAll}, which contains multiple categories.
We do not use coarse data during training, due to our boundary loss which requires fine boundary annotation.

\textbf{Network architecture.} In the paper, the segmentation network we adopted is the DeeplabV3+ (backbone: resnet50 )~\cite{chen2017rethinking}.
Some popular network architectures (Unet~\cite{ronneberger2015u-net:}, FPN~\cite{lin2017feature},
Linknet~\cite{chaurasia2017linknet:}, PSPNet~\cite{zhao2017pyramid}, PAN~\cite{li2018pyramid}) are also tested.

\textbf{Parameter setting.} The parameters are set as follows: $\tau=1, \lambda=10, \xi=1, \zeta=1, \eta=1$.
In the generation of pseudo samples,
 the disk strel of radius $r$ for the dilation and erosion operation is randomly sampled integer between $11$ and $55$.
The interval iteration number between the segmentation network and discriminators is $5$, the batch size is $64$,
Adam hyperparameters for two discriminators $\alpha =0.0001,\beta_{1}=0,\beta_{2}=0.9$. The learning rate for the segmentation network and two discriminators are set as $1e^{-4}$.

\textbf{Metric.} The metrics we adopted include Pixel Accuracy (PA), Mean Pixel Accuracy (MPA),
Mean Intersection over Union (MIoU), and Frequency Weighted Intersection over Union (FWIoU).
Since the Dice index and IoU are positively correlated, the Dice index is omitted.


\begin{table*}[!t]
\centering
\resizebox{\textwidth}{!}{
\begin{tabular}{c c c c c c c c c c}
\toprule
\multicolumn{1}{c}{\textbf{Category}}  &{\textbf{C-Num.$=1$}} & {\textbf{C-Num.$=2$}} & {\textbf{C-Num.$=3$}} & {\textbf{C-Num.$=4$}}  & {\textbf{C-Num.$=5$}} & {\textbf{C-Num.$=6$}} & {\textbf{C-Num.$=7$}} & {\textbf{C-Num.$=8$}} & \textbf{Average}\\
\cmidrule(r){1-1} \cmidrule(r){2-9} \cmidrule(r){10-10}
\emph{bicycle} &   {44.83 / 62.52} & { 42.93 / 56.75} & { 30.85 / 61.01} & { 33.11 / 61.84} & { 35.78 / 59.21} & { 36.52 / 57.25  } & { 36.66 / 55.72  } & { 37.12 / 51.91  } & {37.23 / 58.28  }  \\
\emph{train} &  $\times$  & { 45.67 / 68.03} & { 48.07 / 67.93  } & { 50.68 / 68.47} & { 49.86 / 69.69} & { 51.35 / 69.52} & { 47.20 / 68.37  } & { 53.95 / 66.94  } & {49.54 / 68.42  } \\
\emph{airplane} &  $\times$  &  $\times$ & { 48.67 / 62.21  } & { 50.04 / 63.83  } & { 50.29 / 65.20} & { 53.13 / 64.79} & { 51.80 / 65.35} & { 49.15 / 63.49  } & { 50.51 / 64.15 } \\

\emph{bird} &  $\times$   & $\times$ & $\times$ & { 41.08 / 63.15} & { 39.40 / 63.20} & { 41.74 / 59.58  } & { 42.50 / 60.53  } & { 41.68 / 61.00  } & {41.28 / 61.49 } \\

\emph{person} &  $\times$  & $\times$ & $\times$ & $\times$ & { 32.39 / 72.29} & { 35.89 / 72.62} & { 34.54 / 71.79}  & { 38.08 / 70.12  }& { 35.22 / 71.71 } \\

\emph{cat} &  $\times$  & $\times$ & $\times$ & $\times$  & $\times$  & { 56.74 / 79.52} & { 61.58 / 80.58} & { 61.75 / 74.86  } & { 60.03 / 78.32 } \\

\emph{car} &  $\times$   & $\times$ & $\times$ & $\times$   & $\times$ & $\times$  & { 39.53 / 71.38} & { 42.06 / 71.27  } & { 40.80 / 71.33 } \\

\emph{dog} &  $\times$  & $\times$ & $\times$ &  $\times$  & $\times$ &  $\times$ & $\times$ & { 59.04 / 66.89   } & { 59.04 / 66.89 } \\
\bottomrule
\end{tabular}
}
\caption{The IoU scores of (Ref-Net / DeepLabV3+) with incremental category number.
`C-Num.' denotes the category number of training dataset.
'$\times$' denotes the training dataset without the corresponding category (All scores in \%).}
\label{increase_number}
\end{table*}

\begin{table*}[!t]
\resizebox{\textwidth}{!}{
\centering
\begin{tabular}{ccccccccccccccc}
\toprule
\textbf{Index}$\backslash$\textbf{Ablation}
& $\mathbi{R}_{self}^{-}$
& $\mathbi{R}_{cond}^{-}$
& $\mathbi{R}_{pseu}^{-}$
& $\mathbi{R}_{inner}^{-}$
& $\mathbi{R}_{outer}^{-}$
& $\mathbi{R}_{disc}^{-}$
& $\mathbi{R}_{dice}^{-}$
& $\mathbi{R}(1)$
& $\mathbi{R}(5)$
& $\mathbi{R}(10)$
& $\mathbi{R}(20)$
& $\mathbi{R}(50)$
& $\mathbi{R}(100)$
& $\mathbi{R}(fully)$ \\
\cmidrule(r){1-1} \cmidrule(r){2-8} \cmidrule(r){9-15}
MPA    &88.31 &88.01 &87.34  &87.78  &86.96  &83.23  &65.40 &74.39 &83.38  &88.42  &91.82  &89.92  &92.06  &92.19   \\
MIoU   &66.12 &63.75 &70.61  &61.20  &72.32  &66.51  &45.08 &55.86 &65.11  &74.84  &65.36  &70.43  &67.90  &76.84   \\
FWIoU  &88.90 &90.35 &88.15  &87.69  &86.31  &89.09  &82.88 &84.49 &88.38  &88.78  &91.58  &90.78  &91.70  &90.01   \\
\bottomrule
\end{tabular}
}
\caption{The ablation study of Ref-Net on THUR.
$\mathbi{R}_{self}^{-}$, $\mathbi{R}_{cond}^{-}$, $\mathbi{R}_{pseu}^{-}$, $\mathbi{R}_{inner}^{-}$, $\mathbi{R}_{outer}^{-}$,  $\mathbi{R}_{disc}^{-}$  and $\mathbi{R}_{dice}^{-}$  denote the Ref-Net without boundary-aware self-supervision, condition images, pseudo triplet, inner boundary discriminator, outer boundary discriminator, two discriminators, and supervised loss (dice loss).
$\mathbi{R}(K)$ denotes the Ref-Net with $K$ labeled samples.}
\label{ablation_table}
\end{table*}

\subsection{Comparing with SOTA Methods}

In this section, the Ref-Net is compared with the SOTA methods,
including \emph{unsupervised methods} (CAC~\cite{hsu2018co-attention}, ReDO~\cite{chen2019unsupervised}),
\emph{few-shot methods} (SG-One~\cite{zhang2018sg-one:}, PANet~\cite{wang2019panet:}, SPNet~\cite{xian2019semantic}, CANet~\cite{zhang2019canet:}),
\emph{weakly-/semi-supervised methods} (USSS~\cite{kalluri2019universal}, ALSSS~\cite{hung2018adversarial})
and \emph{fully supervised methods } (boundary-aware methods:\{ Gated-SCNN~\cite{takikawa2019gated-scnn:}, BFP~\cite{ding2019boundary-aware}\}, Unet~\cite{ronneberger2015u-net:}, FPN~\cite{lin2017feature}, LinkNet~\cite{chaurasia2017linknet:}, PSPNet~\cite{zhao2017pyramid}, PAN~\cite{li2018pyramid}
and DeeplabV3+~\cite{chen2017rethinking}) on six datasets.
For the semi-supervised methods, ten labeled samples are provided.
Except for the panoramic, the target dataset and open-source dataset have no overlapped object category.
For a fair comparison, the categories of each multiple category dataset (SBD and THUHR) are split into two non-overlapping parts.
The fully supervised methods are trained with both two parts.
The transfer learning based methods are initially trained on the half-category samples and then trained with specified labeled samples of the rest half-category.
Table~\ref{comparing_SOTA} shows the quantitative results,
where we can see that most scores of $\mathbi{R}(10)$ achieve the state-of-the-art results on par with existing non-fully supervised methods.
Even with more labeled samples,
$\textbf{Trans.}(100)$ only achieves higher scores than $\mathbi{R}(10)$ on THUR and Bird dataset.
Moreover, with only $10$ labeled samples, the  Ref-Net can achieve better results than some fully supervised methods and close results on par with the best fully supervised method.
Meanwhile, with fully supervised samples, the Ref-Net achieves higher scores on the complex dataset (Cityscapes and SBD), which demonstrates the advantage of Ref-Net for handling datasets with more categories.
Note that the resolution of the Cityscapes dataset we adopted is $128*128$. The above two causes lead to the relatively low-scores of all the methods. However, it still validates the superior performance and wide application of the Ref-Net.

\subsection{Results on Incremental Category Number}

Table~\ref{increase_number} gives the IoU scores of Ref-Net and DeepLabV3+ with the incremental category number.
For the Ref-Net, the eight categories are set as target dataset, and ten labeled samples are provided for those categories. The rest of the categories are set as the open-source dataset.
From Table~\ref{increase_number},
we can see that the IoU scores of \emph{bicycle} increase with the incremental category number when the `C-Num.' larger than $3$.
The first two scores ($44.83$ and $56.75$) of the \emph{bicycle} are larger than the score $30.85$ (`C-Num.=3'). The reason is that the network trained with a single category will learn more category-aware features and focus on the single category. When the category number increases, the IoU scores first decrease then increase, which indicates that the Ref-Net can learn more general segmentation ability with the incremental category number.
For the rest seven categories,
most of the IoU scores increase with the incremental category number.
In contrast, most IoU scores of DeeplabV3+ decrease with the incremental category number, which verifies the drawback of model-level knowledge translation.

\subsection{Ablation Study}
To verify each component's effectiveness, we conduct an ablation study on the boundary-aware self-supervision, the pseudo triplet, two boundary discriminators, supervised loss, and the different numbers of labeled samples.
In the experiment, the knowledge translation is set as \{THUR $\leftarrow$ MixAll$^{-}$\}.
For all $\mathbi{R}^{-}$, ten labeled samples of each category are provided.
From Table~\ref{ablation_table}, we can see that $\mathbi{R}(10)$ achieves higher scores than others, which verifies the  effectiveness of each component.
In addition, $\mathbi{R}(10)$ achieves
about $2\%$ increase on the scores of $\mathbi{R}_{inner}^{-}$ and $\mathbi{R}_{outer}^{-}$ and about $5\%$ increase on the scores of $\mathbi{R}_{disc}^{-}$, which demonstrates that the two-discriminator framework is useful for improving the segmentation performance.
For the different numbers of labeled samples, we find that $10$-labeled-samples is a critical cut-off point, which can supply relatively sufficient guidance. With more labeled samples, Ref-Net achieves better performance.

\section{Conclusion}

In this paper, we propose the Ref-Net for segmenting target objects with a reference image as condition,
which is inspired by the human recognition mechanism with a reference object.
The Ref-Net contains two modules: a Reference Segmentation Module (RSM) and a Boundary Knowledge Translation Module (BKTM).
Given a reference object, RSM is trained for finding and segmenting the same category object with tens of finely-grained annotations.
Meanwhile, MMD and boundary-aware self-supervision are introduced to constrain the representation consistency of the same category and boundary consistency of the segmentation mask, respectively.
Furthermore, in BKTM, two boundary discriminators are devised for distinguishing whether the segmented objects and background contain residual features.
BKTM is able to translate the boundary annotation of open-source samples into the segmentation network.
The open-source dataset and target dataset can have totally different object categories.
Exhaustive experiments demonstrate that the Ref-Net achieves close results on par with fully supervised methods on six datasets.
The most important of all, the segmentation performance of all categories will improve with the increasing categories in the target dataset,
while existing methods are opposite. The root reason is that the proposed method is based on dataset-level knowledge translation,
where the data flow of open-source samples will not pass the segmentation network.
It brings a new perspective for the framework design of the image segmentation task.

\section*{Acknowledgments}
This work is supported by National Natural Science Foundation of China (No.62002318), Zhejiang Provincial Natural Science Foundation of China (LQ21F020003), Key Research and Development Program of Zhejiang Province (2020C01023), Zhejiang Provincial Science and Technology Project for Public Welfare (LGF21F020020), Ningbo Natural Science Foundation 202003N4318), Major Scientific Research Project of Zhejiang Lab (No.2019KD0AC01), and Zhejiang Lab (No.2020KD0AA06).

\bibliographystyle{named}
\bibliography{ijcai21}

\end{document}